\documentclass[final]{l4dc2026}

\usepackage{float}
\usepackage{subcaption}
\usepackage{times}
\usepackage{booktabs}
\usepackage{multirow}
\usepackage[table]{xcolor}
\usepackage{pifont}
\usepackage{makecell}
\usepackage{amsmath}
\usepackage{wrapfig}
\newcommand{\cmark}{{\color{green!60!black}\ding{51}}}
\newcommand{\xmark}{{\color{red!70!black}\ding{55}}}

\title[Offline RL for Plasma Control]{Offline Reinforcement Learning for Rotation Profile Control in Tokamaks}

\author{%
  \Name{Rohit Sonker}$^{1}$
  \Name{Hiro Josep Farre Kaga}$^{2}$
  \Name{Jiayu Chen}$^{1,4}$
  \Name{Andrew Rothstein}$^{2}$
  \Name{Ian Char}$^{5}$\\
  \Name{Ricardo Shousha}$^{3}$
  \Name{Egemen Kolemen}$^{2,3}$
  \Name{Jeff Schneider}$^{1}$ \\
  \addr $^{1}$Robotics Institute, Carnegie Mellon University\\
  \addr $^{2}$Princeton University\\
  \addr $^{3}$Princeton Plasma Physics Lab\\
  \addr $^{4}$The University of Hong Kong\\
  \addr $^{5}$Lila Sciences \thanks{work done while at Carnegie Mellon University}
}

\begin{document}

\maketitle

\begin{abstract}%

% \ianx{In my opinion, the abstract still goes into a bit too much detail for a non-fusion audience. I am worried that an ML reader would get overwhelmed by the first few sentences in this abstract.}
Tokamaks remain leading candidates for achieving practical fusion energy, yet many important control problems inside these devices are still difficult or unsolved. One such challenge is controlling the plasma rotation profile, which strongly influences stability, confinement, and transport. While the average rotation can be controlled, controlling the full profile is challenging due to high dimensionality, response to multiple actuators and dependence on plasma condition. Learning-based control methods, such as reinforcement learning (RL), provide a potential solution to this challenging problem with an ability to model complex interactions leading to effective multi-input multi-output control. However, learning such policies is challenging due to the lack of accurate simulators that can model the rotation profile dynamics. In this work, we investigate the use of offline RL and offline model-based RL algorithms for rotation profile control, training them solely on historical data from the DIII-D tokamak. Our final method uses probabilistic models of plasma dynamics to generate rollouts for RL training. We deploy this policy on the DIII-D Tokamak and observe promising real-world results. We conclude by highlighting key challenges and insights from training and deploying an RL policy on a complex physical device while using only limited past data.
% \ianx{What will your learnings be? Are they different from my original paper? It may be hard to have this if it is largely the same. However, I do feel like you may have enough content that this isn't necessary. Especially if you do an evaluation of offline algos on the test-model.}
%
\end{abstract}


\section{Introduction}
% \begin{itemize}
%   \item Limit the main text (not counting acknowledgments or references) to 10 PMLR-formatted pages, using this template.
%   \item Include {\em in the main text} enough details, including proof details, to convince the reviewers of the contribution, novelty and significance of the submissions.
%   \item Make sure to cite references for related literature and for any claims that are not established in the paper using the \texttt{\textbackslash citep} and \texttt{\textbackslash citet} commands, e.g. \citep{levine2016end}, \citet{doyle2003guaranteed}, \citet[Theorem 2.2.1]{vershynin2018high}. We have provided a sample \texttt{.bib} file.
% \end{itemize}
Nuclear fusion has long been regarded as one of the most promising paths toward a stable and abundant energy future. Tokamaks, large toroidal devices where magnetic fields confine a rotating plasma, have emerged as the leading candidate to achieving sustained fusion conditions. Although significant progress has been made in plasma shaping, heating, and current-profile control \citep{ambrosino}, many control problems remain challenging due to nonlinear plasma dynamics and changing operating conditions.

% While progress has been made in plasma shaping, heating, and current-profile control, guided by both theory and decades of experiments \citep{ambrosino}, many challenges remain. The plasma is governed by nonlinear, multi-scale dynamics that are not fully understood, and its behavior often shifts with machine conditions, operating scenarios, and interactions within the magnetic geometry. As a result, developing reliable and high-performance control strategies continues to be a difficult task in tokamak research.

% Tokamak research began in the mid-20th century and has since grown into a global effort, with major facilities operating in the United States, Europe \update{and Asia}. \citep{change_tokamaks}.

\begin{figure}[t]
    \centering
    \includegraphics[width=0.95\linewidth]{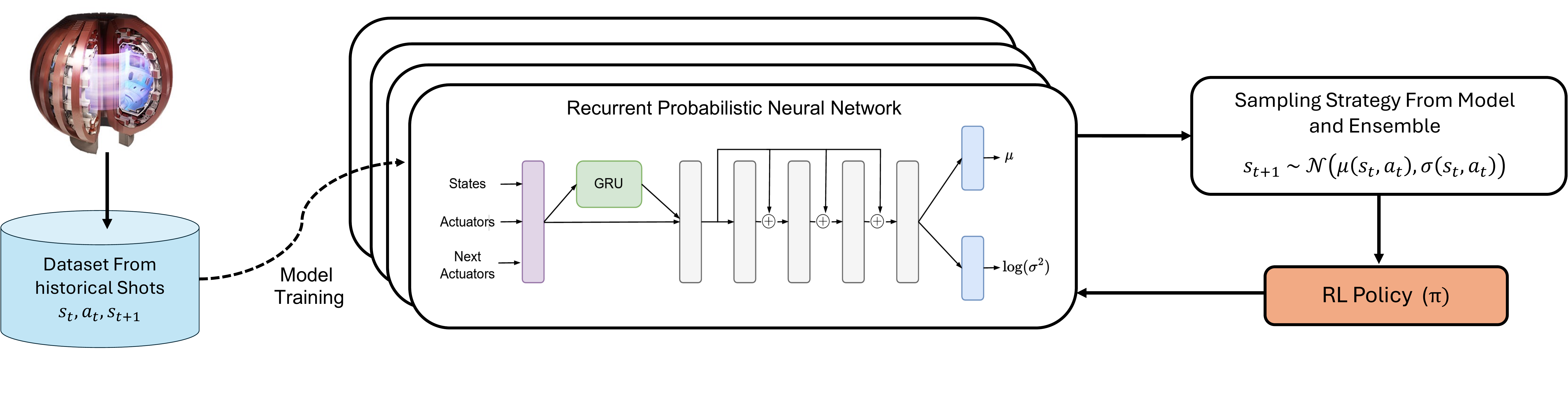}
    \caption{RL Policy Training using trajectories generated autoregressively from the RPNN dynamics model. The model is trained from historical experiment runs at the DIII-D Tokamak.}
    \label{fig:training_loop}
\end{figure}

\begin{figure}[t]
    \centering
    \includegraphics[width=0.95\linewidth]{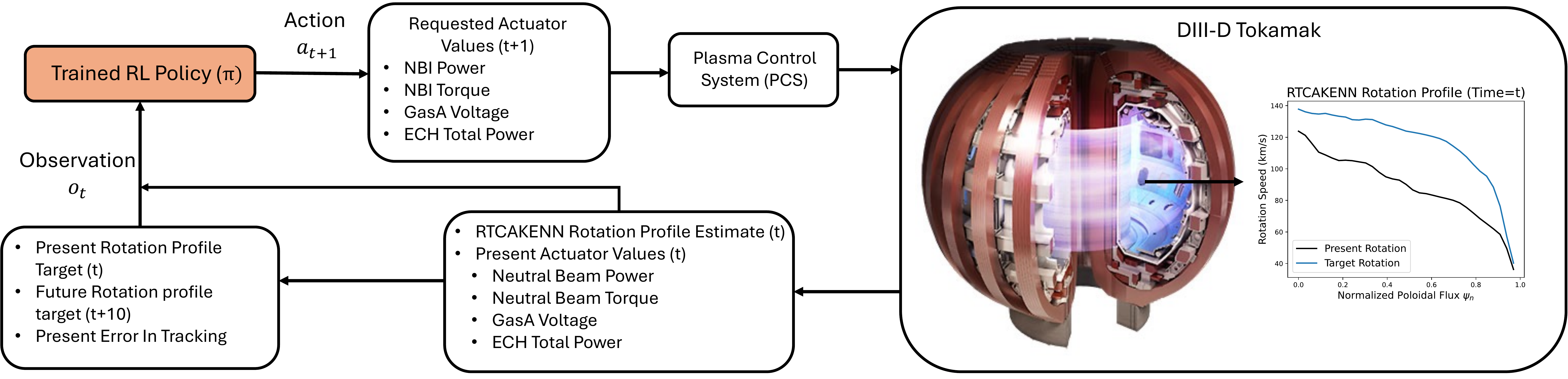}
    \caption{RL Policy Deployment on the DIII-D Tokamak. The policy is uploaded onto the  Plasma Control System (PCS) where it receives signals and real profile estimates from RTCAKENN. The rotation profile variation can be seen from high value at the core to lower values at the edge. Tokamak diagram courtesy of DIII-D.}
    \label{fig:deployment_loop}
\end{figure}

In recent years, developments in Reinforcement Learning (RL) have led to significant progress in many applied domains, where a long horizon optimization problem needs to be solved in complex dynamics. RL has recently been explored for tokamak control, but obtaining on-policy data is difficult because experiments are expensive. Two main approaches have emerged to handle this limitation. \citep{degrave2022magnetic} was the first application of RL to tokamaks for learning shape control. This line of work uses a physics-based simulator during training and then deploys the learned policy directly on the tokamak. A different line of work, for example \citep{char2023offline}, learns a state transition model from historical data and then uses it for RL policy training. These works are promising, yet, they only target a limited set of tasks such as shape control or controlling scalars in the plasma (normalized plasma pressure, or current) for which traditional controllers already exist and are in use on tokamaks \citep{galperti_overview_2024,walker_next-generation_2003,boyer_feedback_2019}.
To realize the full potential of learning-based control, we must tackle tasks where traditional methods are insufficient.

In this work, we address the challenging task of controlling the rotation profile in tokamaks. The profile here refers to the spatial variation in rotation value from core (innermost point) to the edge (outermost point) in the plasma. Controlling the rotation profile is important as it contributes to plasma stability and confinement \citep{wehner2015toroidal}. Traditionally, rotation is controlled as a spatially averaged quantity  by adjusting the neutral beam injection power (NBI or NB power)  and the neutral beam injection torque (NBI or NB torque). Full profile control has been explored in prior works, \citep{wehner2015toroidal, wehner2017combined} show rotation profile control in simulation with MPC. The challenge stems from controlling a high dimensional profile which is affected by interactions from various nonlinear processes and also shows radial coupling - changes at one location impact profile at another location. Traditionally, neutral beam injection power and torque are the main actuators because they directly supply heating and momentum to the plasma. In addition, earlier DIII-D studies showed that Electron Cyclotron Heating (ECH) can modify toroidal rotation \citep{degrassie1999plasma}, and more recent work demonstrated sustainment of differential rotation using feedforward ECH and gas fueling \citep{bardoczi2025tearing}. These prior studies motivate the actuator choices used in this work and show that full rotation profile control in tokamaks is a multi-input, multi-output problem, a setting where simple PID approaches become difficult to apply effectively.

% Hence, full profile rotation control in tokamaks naturally becomes a multi-input, multi-output problem, a setting where classical approaches such as PID control are insufficient. 

% Our goal in this work is to extend past Reinforcement Learning methods to the challenging problem of rotation profile control in Tokamaks. We limit ourselves to approaches that only rely on past data and do not utilize physics based simulators. This direction is based on the analysis by \citet{abbate2024large}, which shows the weakness of these models in providing accurate profile estimates. Another recent work, \citet{wang2025learning} shows that purely physics simulated models may not lead to accurate learning based control. Thus, our problem falls in the setup of offline reinforcement learning, where we only use a static dataset of past experiments to learn policies. This dataset corresponds historical runs from the DIII-D Tokamak,  a device operated by General Atomics in San Diego, California.  

% 

Our goal in this work is to extend past reinforcement learning methods to the challenging problem of rotation profile control in tokamaks. We restrict our approach to methods that rely only on past data and do not utilize physics-based simulators. This choice is motivated by the analysis of \citet{abbate2024large} and \citet{wang2025learning} that show that purely physics-simulated models may not lead to accurate learning-based control. We therefore study this problem in the offline reinforcement learning setting, where policies are learned from a static dataset of past experiments from the DIII-D tokamak, operated by General Atomics in San Diego, California.

Thus, our contribution is to demonstrate the use of offline RL approaches on the task of full profile rotation control, a task which has not been addressed before, using only historical data. Our actuator space utilizes ECH and Gas Puffing in addition to commonly used neutral beams. We use a bootstrapped probabilistic dynamics model ensemble for learning tokamak dynamics (state transition probability), which supports robust policy training and provides an accurate test environment. Under this setup, we benchmark popular offline RL and offline model-based RL algorithms, demonstrating how RL can be applied to such a complex system with limited data and lack of accurate simulations. Finally, we show promising results in deploying an RL policy at DIII-D tokamak facility, while highlighting important lessons and challenges.

\section{Related Works}

\textbf{Reinforcement Learning in Real-World Systems: }Reinforcement learning has been applied across many real-world domains—from legged locomotion \citet{lee2020learning} and robotic manipulation \citet{levine2016end} to control of industrial power systems \citet{su2025review}. 
Advances in stability, expressiveness, and data efficiency have helped RL move closer to practical deployment, yet many challenges remain.
Model-free RL \citet{haarnoja2018soft, schulman2015trust, schulman2017proximal, mnih2015human} often requires large amounts of on-policy interaction, which is rarely available for safety-critical systems. Offline RL \citet{levine2020offline, kumar2020conservative, fujimoto2019off} partly addresses this issue by learning entirely from past data and combining successful trajectories to form a policy. 

\textbf{Model-Based RL and Uncertainty: }Model-based RL offers another path by learning a dynamics model from data and then training a policy using the learnt model as a simulator. These methods can be more data-efficient than model-free approaches \citep{chua2018deep}, but inaccurate or overconfident models may lead to model exploitation, where a policy performs well in simulation but fails on the real system \citep{janner2019trust}. Managing epistemic uncertainty is therefore essential, and probabilistic ensembles have become a standard way to capture uncertainty and stabilize training \citep{chua2018deep}. Model-based offline approaches such as MOPO \citet{yu2020mopo} and MOBILE \citet{DBLP:conf/icml/SunZJLYY23} extend this idea by limiting rollouts to well-modeled regions or penalizing uncertain predictions.

% Model-based RL is closely connected to model-predictive planning methods such as MPPI \citep{williams2017}, the Cross-Entropy Method \citep{botev2013cem, chua2018deep}, and other sampling-based controllers, which evaluate large sets of candidate trajectories through a learned or analytical model. Although these methods have shown strong performance in robotics and continuous-control tasks, they rely on performing hundreds or thousands of model rollouts at every control step, making them unsuitable for tokamak operation. Gradient-based MPC methods \citep{amos2018differentiable} face similar limitations. These approaches differentiate through the dynamics model and optimize a cost function online, requiring repeated forward and backward passes through the model during each control cycle. Again, this is challenging for the tokamak setup, since modelling the nonlinear dynamics requires high capacity complex models. This models have a high inference time; hence, real-time control through planning based methods becomes difficult. Model-based RL avoids these bottlenecks by using the model only during training. Once the policy has been trained, it can be executed directly with a simple forward pass, eliminating the need for online planning or online optimization.

\textbf{Planning-Based Control and its Limits: } Model-based RL is closely connected to model predictive control methods such as MPPI \citet{williams2017}, the Cross-Entropy Method \citet{botev2013cem}, and other sampling-based controllers that evaluate many candidate trajectories through a model. These methods often require hundreds or thousands of rollouts at each control step, which is not feasible for real-time tokamak control, since accurate dynamics models are very complex, and hence have higher inference time. Model-based RL avoids these issues by using the model only during training; once a policy is learned, it runs with a single forward pass and no online optimization.

% Controlling the plasma rotation profile in a tokamak has many advantages. In work by \citet{richner2024use}, differential rotation was shown to affect suppression of tearing instabilities. Differential rotation is the difference in rotation value at two specific points in the profile (these points correspond to locations where the plasma safety factor become 1 and 2 respectively). Furthermore, rotation near the plasma edge (outermost region on profile) affects how neutral gas penetrates the plasma, which can create uneven (asymmetric) fueling in different parts of the tokamak \citep{emdee2024influence, wilkie2024reconstruction}. Therefore, controlling the rotation profile is important for leveraging these effects to improve plasma performance. There has been limited work in controlling the plasma rotation profile. While specific points can be controlled using neutral beam power and torque, controlling the full profile is challenging due interactions from various non linear processes and also radial coupling in the profile. \citet{wehner2015toroidal} presented a linear quadratic optimal controller based on physics model of the dynamics. In a further work, \citet{wehner2017combined} use MPC to control the profile using physics based models. However, both these works are limited to simulation testing. In our experience, the sim2real transfer in deploying controllers at a tokamak is a significant challenge, which we discuss further in this work.

\textbf{Rotation-Profile Control in Tokamaks:} Controlling the plasma rotation profile provides several benefits, differential rotation affects suppression of tearing instabilities \citet{richner2024use} and edge rotation affects neutral-gas penetration, leading to asymmetric fueling \citet{emdee2024influence, wilkie2024reconstruction}.
Because of these effects, rotation-profile control is important for improving plasma performance. However, controlling the full profile is difficult due to nonlinear interactions and radial coupling. Early work by \citet{wehner2015toroidal} proposed a linear–quadratic controller based on physics-based models, and \citet{wehner2017combined} extended this using MPC, though both were limited to simulation studies. In practice, sim-to-real transfer remains a major challenge, which we discuss later in this work.

% In tokamak fusion, learning based methods have been applied for feedforward and feedback control. For feedforward tasks, \citet{Mehta_2024} used bayesian optimizaiton to design safe rampdowns, \citet{sonkermulti} utilized priors from dynamics modelling and used bayesian optimization to optimize for plasma stability. Closed loop reinforcement learning based controls have been developed for various tasks. \citet{degrave2022magnetic} initially showed RL control for plasma shape control at TCV tokamak, which was followed by \citet{TRACEY2024114161} showing further improvement in controlling shape parameters at TCV tokamak. \citet{kerboua2024} followed a similar appraoch to control shape at the WEST tokamak. These methods are able utilize boundary condition simulators since the dynamics for
% the plasma’s shape is more well-understood than other aspects of the plasma, hence can be modeled and controlled better \citep{walker1997multivariable, walker2020introduction}.

\textbf{Learning-based control in Tokamak Fusion:} Learning-based methods have been applied to several feedforward and feedback control tasks in tokamaks. For feedforward control, \citet{Mehta_2024} used Bayesian optimization to design safe rampdowns, and \citet{sonkermulti} used Bayesian optimization with dynamics-informed priors to improve stability. Closed-loop RL has been used for plasma shape control at the TCV Tokamak, \citet{degrave2022magnetic} \citet{TRACEY2024114161} and at the WEST tokamak \citet{kerboua2024}. These works rely on boundary-condition simulators because plasma shape dynamics are relatively well understood and easier to model \citet{walker1997multivariable, walker2020introduction}. Recent progress in data-driven plasma modeling \citet{abbate2021data} and \citet{char2024full} has enabled new learning-based control approaches. \citet{char2023offline} used such models to train RL controllers for normalized pressure and differential rotation. \citet{seo2024avoiding} developed policies that maintain high pressure while avoiding tearing modes. \citet{wu2025high} applied model-based RL to control current and shape parameters. Most recently, \citet{wang2025learning} combined physics priors with historical data to model plasma dynamics and design RL rampdown trajectories for the TCV tokamak. However, in this work, we rely solely on historical data and learn RL policies for rotation-profile control directly from past experiments.

% Significant progress has been made in terms of data-driven dynamics modeling of plasma. \citet{abbate2021data} and \citet{char2024full} have developed data-driven dynamics models for plasma evolution. \citet{char2023offline} also used data-driven models to learn RL policies for normalized plasma pressure and the differential rotation scalar. \citet{seo2024avoiding} developed policies for high plasma pressure while avoiding tearing modes. Whereas \citet{wu2025high} also used a model-based RL policy to control current and plasma shape parameters. Recently, \citet{wang2025learning} 
% presented an approach that combined physics priors and historical data to model plasma dynamics. They then used RL to design ramp down policies for the TCV tokamak. In this work, we only utilize data driven models and learn RL policies for rotation profile control only using the past data.

\section{Methodology}

\subsection{Problem Setup}
% We pose tokamak profile control as a discrete-time, Markov decision process (MDP). In particular we define, $\mathcal{M} := (\mathcal{S}, \mathcal{A},\mathcal{T}, \gamma, R, d_0)$ . $\mathcal{S}$ describes the full plasma state, $\mathcal{A}$ denotes the action space and $\mathcal{T} : \mathcal{S} \times \mathcal{A} \rightarrow \mathcal{P}(\mathcal{S})$ denotes the transition function, where $\mathcal{P(\mathcal{S})}$ denotes the space of probability distributions over $\mathcal{S}$. Each timestep respresents a transition of 20ms. $\gamma$ denotes the discount factor, $R : \mathcal{S}\times\mathcal{A} \rightarrow \mathbb{R}$ denotes the reward function and $d_0$ denotes the initial state distribution. Observing the whole plasma state is difficult in realtime due to diagnostic constraints, hence we learn a policy for the Partially observable MDP, where $\mathcal{O} : \mathcal{O} \subset \mathcal{S}$ denotes the limited observation space of the policy. Thus, the RL goal is to learn policy $\pi : \mathcal{O} \rightarrow \mathcal{P}(\mathcal{A})$, which maximizes the expected sum of discounted rewards $\mathbb{E} [\sum_{t} \gamma^t R(s_t,a_t)]$, where $s_0 \sim d_0, a_t \sim \pi(o_t)$, and $s_{t+1} \sim T(s_t, a_t)$. A list of variables that we use for the State, Action and Observation space are provided in table \ref{tab:signals

We frame tokamak profile control as a discrete-time Markov decision process (MDP) $\mathcal{M} = (\mathcal{S}, \\ \mathcal{A}, \mathcal{T}, \gamma, R, d_0)$. Here, $\mathcal{S}$ is the plasma state, 
$\mathcal{A}$ is the action space, and $\mathcal{T} : \mathcal{S} \times \mathcal{A} \rightarrow \mathcal{P}(\mathcal{S})$ is the transition distribution over next states. Each timestep corresponds to 20 ms. The reward function is $R : \mathcal{S} \times \mathcal{A} \rightarrow \mathbb{R}$, $\gamma$ is the discount factor, and $d_0$ is the initial-state distribution. Because the full plasma state cannot be observed in real time due to diagnostic limits, we instead learn a policy for a partially observable MDP. Let $\mathcal{O} \subset \mathcal{S}$ denote the observation space available to the policy. The objective is to learn a policy $\pi : \mathcal{O} \rightarrow \mathcal{P}(\mathcal{A})$ that maximizes the expected discounted return
$\mathbb{E}\left[\sum_t \gamma^t R(s_t, a_t)\right], \quad
s_0 \sim d_0,\; a_t \sim \pi(o_t),\; s_{t+1} \sim \mathcal{T}(s_t, a_t)$.
The specific state, action, and observation variables used in this work are listed in Table~\ref{tab:signals}

\subsection{Dynamics Model Learning}
 
% We first aim to learn the dynamics of the plasma such given a current state and action value, we get a distribution over the next state $s_{t+1} \sim T(s_t, a_t)$. For this we choose the recurrent probabilistic neural network (RPNN) \cite{char2024full}, as our dynamics model, which predicts the parameters of a guassian distribution (mean and log of variance) i.e. we learn to predict $s_{t+1} \sim \mathcal{N} (\mu_{t+1}, \sigma_{t+1})$ where for every timestep, $\mu_{t+1}, \sigma_{t+1} = T_\theta (s_t, a_t)$ where $T_\theta$ is an RPNN model parameterized by $\theta$. To predict these quantities the RPNN has a mean $\mu$ head and $\log (\sigma^2)$ head. A diagram of the network architecture can be seen in Fig \ref{fig:training_loop}.
We first aim to learn the plasma dynamics, such that given a state–action pair $(s_t, a_t)$, we obtain a distribution over the next state $s_{t+1} \sim T(s_t, a_t)$.  For this, we use the recurrent probabilistic neural network (RPNN) of \citet{char2024full} as our dynamics model. The RPNN predicts the parameters of a Gaussian distribution (mean and log variance), such that $s_{t+1} \sim \mathcal{N} (\mu_{t+1}, \sigma_{t+1})$ where for every timestep, $\mu_{t+1}, \sigma_{t+1} = T_\theta (s_t, a_t)$ where $T_\theta$ is an RPNN model parameterized by $\theta$. For this, the RPNN has a mean head and a log variance head. A diagram of the network architecture can be seen in Fig.\ref{fig:training_loop}.

 Our model is trained on roughly 18,000 past DIII-D experiments (“plasma shots”). Each shot contains about four seconds of data sampled at 20 ms. We restrict training to the flat-top phase of the discharge, where the plasma is held in steady, high-performance conditions. The full list of states and actions is given in Table~\ref{tab:signals}. For profiles i.e., electron temperature, ion temperature, pressure, rotation, and the q profile, we use Zipfit reconstructions \citep{logan2018omfit}, which provide smooth, physically constrained estimates. These reconstructions are not available in real time, so during deployment the policy instead uses RTCAKENN predictions \citep{shousha2023machine}. This creates a noticeable sim2real gap, which we account for later in the policy-learning and discussion sections. We tried training a model with RTCAKENN data, however the model did not learn well, due to large noise and variations in the signal. Following \citet{char2024full}, we reduce all profile quantities using PCA: 4 components each for electron temperature, ion temperature, and rotation, and 2 components each for pressure and the q profile, which captures more than 99\% of the variance.

% We make a couple changes to the RPNN training process which leads to improved model performance. First, we find that training the RPNN in two stages - stage 1 on mean squared error (MSE) loss and stage 2 where only the $\log(\sigma^2)$ head of the network (see fig \ref{fig:training_loop}) is trained on negative log likelihood (NLL) loss, leads to better explained variance scores. Moreover, we train an ensemble of RPNN models via bootstrapping on the dataset. While we get a measure the alleatoric uncertainty from the $\log(\sigma^2)$ prediction of the RPNN, the disagreement between the bootstrapped ensemble members provides us with an estimate of the epistemic uncertainty \citep{chua2018deep}. More details on the RPNN training are provided in the Appendix.
We introduce two modifications to the RPNN training procedure that improve performance. First, we train the model in two stages: (1) an initial phase using mean squared error loss, and (2) a second phase where only the log variance head is trained using negative log-likelihood. This leads to higher explained-variance scores. Second, we train a bootstrapped ensemble of RPNNs. The predicted log variance captures the aleatoric uncertainty, while disagreement across ensemble members provides an estimate of epistemic uncertainty \citep{chua2018deep}. Further training details are given in the Appendix. This ensemble of probabilistic models serves as the evaluation environment for all policy-learning methods, since direct testing on the tokamak is prohibitively expensive. 
% For offline model-based RL algorithms, we also use this dynamics model with a different sampling strategy, described in the next section.

% This ensemble of probabilistic models, provides us a with a testing environment for all policy learning methods, since evaluating them on the real tokamak is exceedingly expensive and  impractical. For offline model based RL algorithms, we also use this dynamics model under a different sampling strategy, which is explained in the next section.

\subsection{RL Policy Learning}

% We now use the learned dynamics model to train an RL policy. To simulate a tokamak experiment run or a ‘shot’, we perform autoregressive rollouts from the historical data. Since we model the tokamak with an ensemble $T_{\theta_i}$, where $i=1,..,25$. Each model $T_{\theta_i}$ provides a plausible dynamics model, and we assume that true device dynamics lies within the distributions spanned by these individual models.
We now use the learned dynamics model to train an RL policy. To simulate a tokamak experiment, or a “shot,” we generate autoregressive rollouts using the ensemble of dynamics models $T_{\theta_i}$, where $i=1,..,25$. Each model in the ensemble represents a plausible version of the device dynamics, and we assume that the true tokamak dynamics lies within the distribution spanned by these models.

% During RL training, we sample one model $T_{\theta_i}$ for each trajectory and generate autoregressive trajectories by sampling from that model’s predictive distribution (mean and log-variance). We resample the model when a new trajectory is drawn. This procedure trains the policy to perform well on average across all plausible dynamics, and because high-variance regions induce larger state dispersion across models, it implicitly reduces synthetic experience in these poorly known regions. Thus, we do not add any explicit uncertainty penalty in the reward.
During RL training, we sample one model $T_{\theta_i}$ for each trajectory and generate autoregressive trajectories by sampling from that model’s predictive distribution (mean and log-variance). A new model is selected when a new trajectory is sampled. This trains the policy to perform well across all plausible dynamics, and because regions of high model variance produce more spread in the sampled next states, the policy naturally receives fewer synthetic transitions in uncertain areas. As a result, we do not include any explicit uncertainty penalty in the reward.

% \textbf{Policy Observation Space} - The policy observation space consists of the current actuator values (Power, Torque, Gas Puffing Voltage and ECH total power). This allows policy to observe the actuator response of the machine. We also use the RTCAKENN of the current rotation profile, and we add the current target profile and the future target profile 10-timesteps ahead to the observation. Note that these profiles (current values and targets) and converted into PCA components before being provided to the policy. Finally, we also compute the current tracking error and add that to the observation space.
\textbf{Policy Observation Space} - The policy observes the current actuator values (neutral beam power, torque, gas-puffing voltage, and total ECH power), which together describe the machine’s present actuation state. We also include the RTCAKENN estimate of the present rotation profile, along with the present target profile and a future target profile shifted 10 timesteps ahead. All profile quantities, i.e., present values and targets, are converted into their PCA components before being passed to the policy. Finally, we also compute the present tracking error and include it as part of the observation space.
%\citet{TRACEY2024114161}

% \textbf{Policy Action Space} - \rohit{Check all physics, princeton people please help!} In this experiment, we use NBI Power, NBI torque , Gas Puffing voltage, and ECH (Electron Cyclotron Heating) total power as the action space. NBI Torque has direct effect on rotation profile at locations where the neutral beams points at, which is generally a point near the plasma core. Since torque and power are provided by the same source, and Neutral beams are responsible for overall plasma maintenance, we also add power as an actuator. We operate in a counter beam configuration, which means that we have independent control of power and torque. Gas Puffing and ECH have more complex relationships with rotation. Gas adds to the density of the plasma, thereby slowing down rotation due to momentum conservation. Similarly, ECH has a negative effect on density and thus leads to an increase in rotation. This set of actuators allows sufficient room to alter the rotation profile.
\textbf{Policy Action Space} - The policy controls NBI power, NBI torque, gas-puffing voltage, and total ECH power. NBI torque directly affects the rotation profile at the beam deposition location, which is typically near the plasma core. Because torque and power originate from the same source, and the neutral beams are central to overall plasma sustainment, we include both as independent actuators.
% We operate in a counter-beam configuration, which gives us independent control over power and torque. 
Gas puffing and ECH influence rotation in more indirect ways: gas particles can increase density, or adding slow moving particles slows the rotation through momentum conservation, starting at the outermost spatial locations and propagating inwards. Whereas ECH tends to reduce density and thereby increases rotation, typically starting at intermediate spatial locations. Together, these actuators provide enough flexibility to meaningfully shape the rotation profile.

% \textbf{Target Sampling}\label{sec:training_target_sampling} - Since plasma experiments are done in various different configurations, it is quite possible that some targets are unachievable in certain experiment setup. Hence, to account for this, we use the following strategy to sample a target during policy training. When replaying a reference shot from the dataset, we pick another \lq target\rq \ shot from the same experiment session as the reference shot. From this target shot, we sample rotation profiles at two time points. Based on these two fixed profiles, we recreate a step function. This process ensures that we only sample profiles that are achievable by the current scenario. 

\textbf{Target Sampling}\label{sec:training_target_sampling} - Because plasma experiments operate in many configurations, some target profiles may be unattainable in a given setup. To address this, we use the following target-sampling strategy during policy training. When replaying a reference shot from the dataset, we select another “target’’ shot from the same experimental session. From this target shot, we sample rotation profiles at two time points and use these two profiles to construct a step-function target. This ensures that all sampled targets are achievable within the operating conditions of the reference scenario, while maintaining target diversity.

\textbf{Reward Function} - We use the negative mean squared tracking error on tracking the whole rotation profile (33 dimensions per timestep), which is reconstructed from the PCA components. 

$$
R(t) = -\left \Vert rot_{target} (t) - rot(t) \right \Vert _2 ^2
$$
where $rot_{target}(t)$ is target rotation at $t$ and $rot(t)$ is the actual rotation at time $t$. 
% Both are 33 dimensions each.

\textbf{Sim2Real Gap and domain randomization} - Although sampling across the dynamics ensemble already provides a form of domain randomization, we further add small observation noise during policy training to improve robustness. We increase the noise level until performance in the test environment begins to degrade. In practice, we add zero-mean Gaussian noise with a standard deviation of 0.1 to the observation vector. 
% \ianx{Might want to consider adding a citation here}

\textbf{Policy Testing} - Since the true test environment is unavailable and tokamak time is limited, we evaluate the policy using the learned dynamics model with a different sampling strategy. Here, the goal is to maximize accuracy relative to the collected data. At each timestep, we sample predictions from all models in the ensemble and take their mean to estimate the next state. Empirically, this strategy provides the highest accuracy on the dynamics-model test set. %The shots used for testing are also held out during the model and policy training.

\section{Results and Discussion}
\subsection{Benchmarking Offline RL Methods}
We first present tracking results using repeated rollouts on three shots that share similar conditions to those used on the experiment day. At the tokamak facility, each experiment relies on a past shot—called the reference shot—to provide stable and well-understood operating conditions. For our experiment, the reference shot was 161409, so we evaluate on this shot along with 161410 and 161412, which were performed in the same session and follow nearly identical setups. This provides the closest reconstruction of the real experiment configuration available from our dataset.

We perform a comparative analysis of various offline RL algorithms using an open-source offline RL codebase tailored for plasma control \citep{OfflineRLKit2025}. Notably, two model-based offline RL algorithms featured in this codebase, BAMCTS \citep{DBLP:journals/corr/abs-2410-11234} and ROMBRL \citep{DBLP:journals/corr/abs-2505-13709}, have been evaluated for rotation profile control in their original studies, but only in simulation. The results in table \ref{tab:offline-rl-reduced} show the tracking accuracy of the full rotation profile in each of these shots with different types of algorithms. A detailed table showing tracking performance at different profile locations is also shown in table \ref{tab:rotation-tracking-full}.

\begin{wraptable}{r}{0.65\textwidth}
\vspace{-8pt} % optional tightening
% \begin{table}
\centering
\caption{Performance of different algorithms on Rotation Profile tracking. Root mean squared error (RMSE), values (lower is better), with standard error in parenthesis, are given for simulated runs of shots 161409, 161410, 161412 with 10 seeds each.}
\label{tab:offline-rl-reduced}
\setlength{\tabcolsep}{2pt}
\renewcommand{\arraystretch}{1.0}
% \begin{tabular}{p{2.3cm} p{5cm} p{2.2cm}}
\begin{tabular}{lcc}
\toprule
\textbf{Category} & \textbf{Algorithm} & \textbf{RMSE} \\
\midrule

\multirow{2}{*}{\makecell[l]{Behaviour\\Cloning}}
 & GCIL \citep{ding2019goal} & 36.48 (0.37) \\
 & & \\
\midrule

\multirow{3}{*}{\makecell[l]{Offline\\Model-Free\\RL}}
 & CQL \citep{kumar2020conservative}  & 80.79 (0.25) \\
 & IQL \citep{kostrikov2021offline}  & 54.52 (0.78) \\
 & TD3BC \citep{fujimoto2021minimalist} & 58.73 (0.51) \\
 & EDAC \citep{an2021uncertainty} & 32.07 (0.66)\\
 & MCQ \citep{lyu2022mildly} & 44.64 (0.44)\\
\midrule

\multirow{5}{*}{\makecell[l]{Offline\\Model-Based\\RL}}
 & COMBO \citep{yu2021combo}  & 51.46 (0.62) \\
 & BAMCTS \citep{DBLP:journals/corr/abs-2410-11234} & 57.43 (0.71) \\
 & ROMBRL \citep{DBLP:journals/corr/abs-2505-13709}& 91.44 (0.70) \\
 & RAMBO  \citep{rigter2022rambo}& 40.05 (0.43) \\
 & MOPO \citep{yu2020mopo} & 84.63 (0.72) \\
 & MOBILE \citep{sun2023model}& 32.99 (0.40)\\
 % & \makecell[l]{\textbf{PPO (with ensemble} \\ \textbf{of probabilistic}\\ \textbf{dynamics models)}} & \textbf{29.50 (0.46)}\\
& \textbf{PPO} \citep{schulman2017proximal} & \textbf{29.50 (0.46)}\\
\bottomrule
\end{tabular}
\end{wraptable}

Goal-Conditioned Imitation Learning (GCIL) and all offline model-free RL methods are trained directly on the same offline dataset used to learn our dynamics models. For offline model-based methods, the learned dynamics model is available under the training sampling mode described earlier. We also evaluate Proximal Policy Optimization (PPO), a model-free method with respect to policy optimization, but we train it with our ensemble of probabilistic dynamics models serving as the simulator. 
% Because the ensemble already captures uncertainty, we do not add additional reward penalties, in contrast to many offline model-based approaches. 
It is also important to note that our dataset is relatively small and corresponds to a highly complex control task. This makes this setting particularly challenging for offline RL methods. Adding conservative penalties in this setting leads to overly cautious behavior, and in practice PPO outperforms these methods. Based on these observations, we select PPO as the final algorithm for deployment.

\subsection{Experiment Results}

\begin{figure}[t]
  \centering
  \includegraphics[width=1\linewidth]{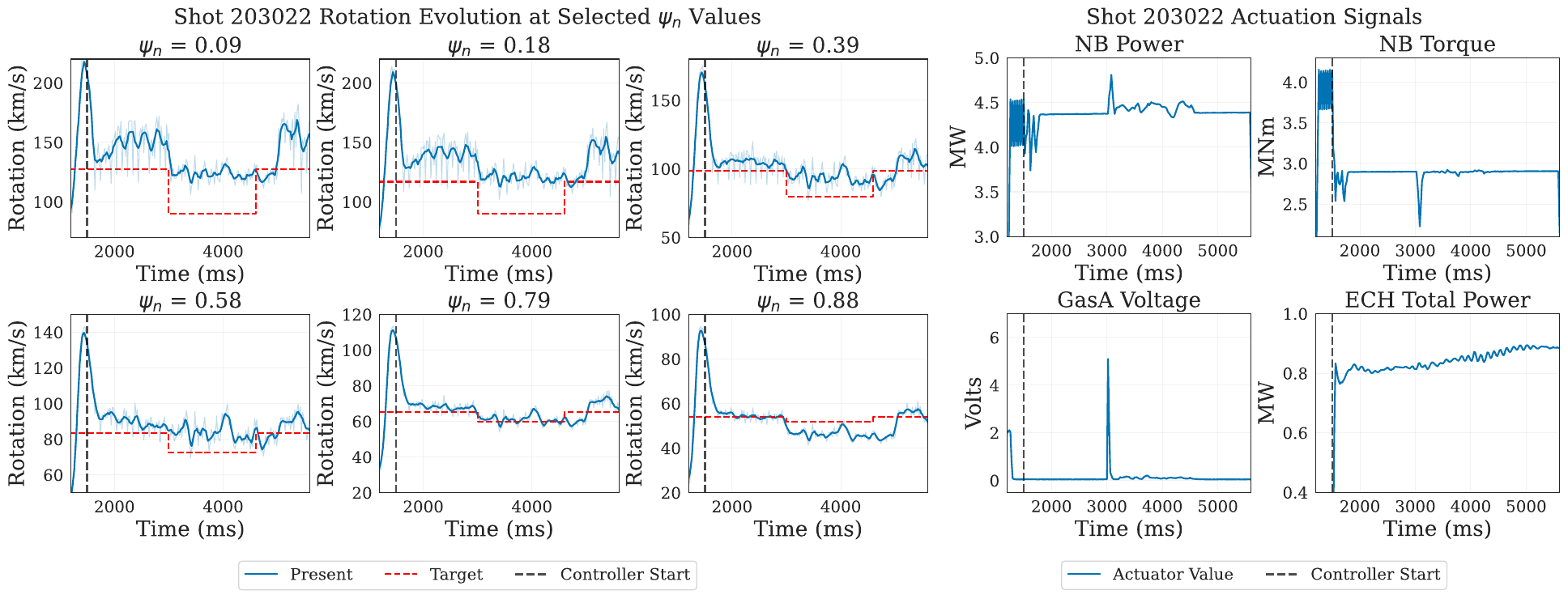}
  \caption{Real tokamak results for Shot 203022 from DIII-D. The first set of plots show value of rotation at different normalized poloidal flux values ($\psi_n$), such that $\psi_n = 0$ represents plasma at the core and $\psi_n=1$ represents plasma at the outer edge. Blue represents the present value while red represents the target. Smoothened lines are shown in dark. Second set of plots shows the changes in actuator signals made by RL policy. Controller starts running at 1500ms when the flat-top phase of the experiment starts. }
  \label{fig:exp_results}
\end{figure}

% We now show the results of online testing at the DIII-D tokamak. Once we have trained our final policy, we deploy it on the Plasma Control System (PCS) at DIII-D. The overall control flow during deployment in shown in fig.\ref{fig:deployment_loop}. Due to challenging experiment conditions and lack of time, we only managed a single shot in the facility. The shot, as shown in fig \ref{fig:exp_results}, shows tracking performance of the algorithm at different $\psi_n$ (normalized flux radius) values. This value, in simple terms, represents how far a point lies between the center of the plasma and its outer edge. 
We now present the results of online testing at the DIII-D tokamak. After training our final policy, we deployed it on the Plasma Control System (PCS) at DIII-D. Keras2c package \citep{keras2c} was used to convert the policy to C code. The full deployment loop is shown in Fig.\ref{fig:deployment_loop}. The tracking performance for this shot is shown in Fig.\ref{fig:exp_results} across several values of $\psi_n$, the normalized flux radius, which indicates how far a point is from the plasma center toward the edge.

% We create a target for the rotation profile by defining a step function with 2 changes as shown in fig. \ref{fig:exp_results}. The RL controller starts at 1500ms once the ramp-up phase of the experiment is complete. The control immediately adjusts the torque to reach the initial target value. Once it reaches 3000ms and the target shifts, we see small increase in power and small decrease in torque, with some addition in gas. This leads to decreases in the rotation profile. Decrease in torque directly impacts the rotation value at the core, while addition of gas impacts the edge first (where it is released), but its effect reaches core as the gas moves in, slowing the rotation. Finally, at 4500ms, the target shifts back to initial setpoint again, power and torque are fluctuations are returned to original baseline. ECH power does not seem to be affected by the middle variation in targets and only follows a slow increase. Thus, even though the tracking performance has an offset, the controller responds well to the step changes with changes across actuators.
For this experiment, we defined a step-function target with two changes, as illustrated in Fig.~\ref{fig:exp_results}. The RL controller was activated at 1.5s, once the ramp-up phase had ended. At this point, the controller quickly adjusted the torque to bring the rotation profile toward the initial target. When the target changed at 3s, we observed a small increase in NB power, a slight reduction in torque, and a sharp rise in gas puffing. These changes jointly produced a decrease in the rotation profile. The decrease in torque directly influences rotation. 
The effect of gas injection is more complex. Around 3s, the gas valve voltage increases sharply, resulting in a higher gas flow into the plasma. Although the injected gas alone is not sufficient to produce the observed change in rotation, it can influence the profile because the newly introduced particles initially have lower momentum. While NB torque does not show a sustained change, we do notice the rotation profile to have changed significantly. This likely involves a nonlinear transient combined with a hysteresis effect that locks the rotation. Interestingly, the controller recognizes the improved state and maintains the actuator setting to sustain it. This highlights the advantage of RL in adapting to evolving plasma conditions.

% The dominant contribution, however, arises from the adjustments in power and torque.

% At 4500 ms, when the target returned to its original value, the controller reversed these adjustments, bringing power and torque back toward their earlier baseline. ECH power did not strongly react to the intermediate target change and instead showed a gradual upward trend throughout the shot. 
At 4.5s, the target changes to a value that closely matches the existing rotation profile, with only a small increase required near the mid-radius region ($\psi_n \approx 0.5$). At this stage, the power and torque adjustments from the previous phase stop, indicating that the controller considers the profile to be close to the desired state. While this causes a small increase in rotation, a pronounced jump appears near the plasma core. We examined possible causes, including tearing instabilities and major sawtooth events, but found no clear evidence of either, and the cause of the spike observed is unclear. 
% Large signal noise is also observed during this time, suggesting that the rise may partly result from artifacts in the RTCAKENN signal.\rohit{reword this, first see if raw values give a jump in rotation} 
At approximately 5.2s, the rotation subsequently moves back toward the target value without any controller contribution. Full profile views are provided in Fig.~\ref{fig:full_prof1} and Fig. ~\ref{fig:full_prof2}.

Overall, while there is a consistent offset in the tracking values, the controller responded to target changes by coordinating the actuators in a physically meaningful way. 

\subsection{Simulated Experiment Results}

\begin{figure}[t]
  \centering
  \includegraphics[width=1\linewidth]{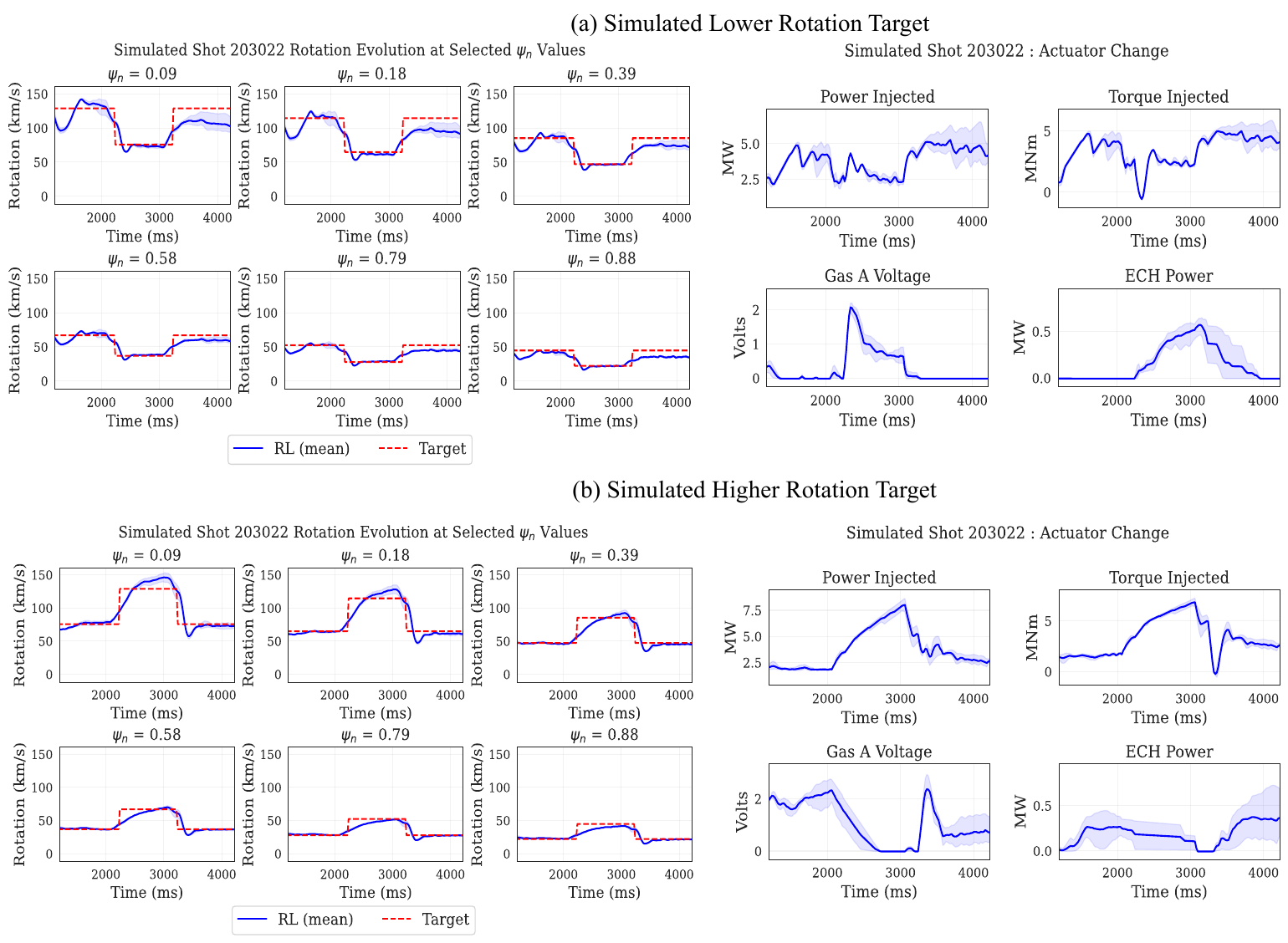}
  \caption{Simulated results for shot 203022 using the dynamics-model environment. Two target patterns are tested: (a) decreasing the rotation profile and returning, matching the experiment, and (b) increasing the profile and returning. Left plots show rotation at different normalized flux values $(\psi_n)$, where $\psi_n=0$ is the core and $\psi_n=1$ the edge (blue: actual, red: target; dark lines: smoothed). Right plots show the RL-controlled actuator signals. Shaded regions denote the 5th–95th percentile over 30 seeds, with the solid line showing the mean. Both cases demonstrate strong tracking performance in the absence of the sim2real gap.}
  \label{fig:sim_results}
\end{figure}

% To further analyze the experiment results and to understand the impact of sim2real gap, we also do a simulated run of the experiment shot in our test environment with the dynamics model ensemble. The major difference lies in the source of input values. Rotation profiles in our simulated environment are based on Zipfit estimates, an offline only algorithm which takes diagnostic measurements and performs functional fitting to produce smooth continuous profiles. However, during the experiment, the policy observes RTCAKENN data. RTCAKENN is neural network based realtime approximation of CAKE (Consistent Automatic Kinetic Equilibrium) data which provides realtime profile estimates. 
To further analyze the experiment results and understand the effect of the sim2real gap, we also simulate the experiment shot using our test environment and the dynamics-model ensemble. The main difference between the real and simulated settings lies in the source of the rotation-profile inputs.
% In simulation, the profiles come from Zipfit estimates—an offline algorithm that processes diagnostic measurements and fits smooth, physically constrained profiles. During the real experiment, however, the policy observes outputs from RTCAKENN, a method which can operate under limited diagnostics and provides real-time but less smooth profile estimates.
In simulation, Zipfit profile (offline profile fitting algorithm) estimates are used, whereas, in realtime, RTCAKENN profile estimates are used.
We run the simulated experiment using two sets of target profiles:
(a) targets that go from a higher rotation level to a lower level and back—matching the pattern used in the actual experiment, and
(b) targets that go from lower rotation to higher rotation and back.
The results are shown in Fig.~\ref{fig:sim_results}.

% The simulated experiment is run with 2 sets of targets - (a) Targets that go from higher rotation to lower and back. This is very similar to targets in the actual experiment. (b) Targets that go from lower to higher and back. The results can be seen in fig. \ref{fig:sim_results}. 

% Lets look at results from Part(a). This is quite similar to the experiment, except that zipfits signals may not always match upto RTCAKENN signals. We notice much better tracking performance by the policy. Highly processed zipfit signals are much more smooth compared to realtime RTCAKENN estimates and this can be seen in the rotation values. Since there are no actuator lags or delays in our model, we see clear changes to support tracking. When the first target change occurs at 2250ms, torque is decreased and we see the corresponding decrease in rotation. We see more decreases in torque and addition of gas to maintain the low rotation value. At the same time, we also see ECH being used to balance the rotation. When the next change occurs at 3250ms, torque and power are both increased, the gas is killed and ECH peaks, subsequently reducing. All these changes lead to an increase in rotation across the profile.
For case (a), the actuator behavior directionally resembles that observed in the real experiment, although major differences exist in magnitudes. These differences are mainly attributed to the source signals used for the profiles, namely Zipfit in simulation and RTCAKENN in the experiment. The RTCAKENN signals are realtime signals used by the experimental policies which are noisy and exhibit high-frequency variations. This strongly impacts the control policy. Tracking performance is noticeably better in simulation because Zipfit profiles are smoother, and, since the dynamics model has no actuator delays or lags, the effects of control actions appear clearly. When the first target drop occurs at 2250 ms, torque decreases and the rotation drops correspondingly. Additional reductions in torque and increased gas puffing help maintain the lower rotation value. ECH is also used to stabilize the profile. When the target increases again at 3250 ms, torque and power rise, gas puffing is shut off, and ECH peaks before tapering off, all contributing to a rise in rotation across the profile. 

% We now look at Part(b) where keep go from a lower rotation to higher rotation and back. Again, we notice that when the target increases, torque is increased, while gas is decreased. This leads to faster rotation. Similarly, when the rotation target is decreased, the torque declines sharply and gas is increased, which leads to drop in rotation. To maintain at that level ECH is also introduced in the last stage. 
For case (b), where rotation moves from low to high and back, we observe similar coordinated behavior. When the target increases, torque increases and gas puffing decreases, leading to faster rotation. When the target drops again, torque is sharply reduced and gas is increased, bringing the rotation down. In the final stage, ECH is introduced to help maintain the lower rotation level.

% This conveys that the RL policy performs well in the simulation and is able to utilize all actuators well to track the complete rotation profile. On the tokamak, however, the policy is quite affected by the sim2real gap. We notice that the targets are tracked but steady state errors exist. The policy makes intelligent decisions in actuator control to control the full profile. 
In summary, the RL policy performs well in simulation and uses all actuators to track the full profile. On the tokamak, however, the policy is affected by the sim2real gap: the targets are followed, but steady-state errors persist. Even so, the actuator responses in the experiment show that the policy makes intelligent and coordinated adjustments, demonstrating its ability to control the complete rotation profile under real operating conditions.

\section{Discussion and Conclusion}

% In this work, we present how offline RL methods (model free and model based) can be utilized for rotation profile control in tokamaks. Our system development and experiment revealed several challenges and insights that point to promising directions for future work. We highlight our key takeaways and learnings for the reinforcement learning community below - 
In this work, we show how offline RL methods—both model-free and model-based—can be used for rotation-profile control in tokamaks. The system development and experiment revealed several challenges that point toward useful directions for the RL community.

% \textbf{Sim2real Gap and RL with Privileged Information: } While sim2real is common for all applications of Offline RL, tokamak setup have a larger distributional shift in going from Zipfit signals to RTCAKENN. Ziptfit is an offline fitting algorithm, whereas RTCAKENN is a real-time prediction of CAKE data which is another profile fitting algorithm. We had limited access to CAKE data hence, could not train dynamics models on it. RTCAKENN data was available however, dynamics modelling failed to perform well on this data. Zipfit data, being smooth due to functional fitting and available in the largest amount, performed much better in profile prediction. Thus, this leads to large sim2real gap. We believe that methods to account for this will lead to significant improvement in performance. This setup also aligns well to the paradigm of RL with priviliged information. Zipfits are computed offline with better information sources, whereas RTCAKENN provides realtime estimates, which are more noisy. Hence, we can utilize RL techniques that can incorporate offline processed profile signals (Zipfits) during training as priviliged information and then work with reduced diagnostics based signals (RTCAKENN) during training. Such techniques have shown  success in robotics \citep{kumar2021rma, lee2020learning} and would be a promising future direction.
\textbf{Sim2Real Gap and RL with Privileged Information.}
Tokamak control has a strong sim2real gap due to the shift from Zipfit signals (offline and smooth) to RTCAKENN signals (real-time and noisier). This mismatch can limit policy performance and suggests that RL methods using privileged information may be useful. Here, Zipfit profiles provide high-quality offline information, while RTCAKENN provides the restricted real-time input. Similar approaches that train with rich states and deploy with limited observations have been successful in robotics \citep{kumar2021rma, lee2020learning} and could be useful here.
\textbf{Actuator Failures.}
During the experiment session, we encountered problems with gyrotrons (ECH) and neutral-beam responses, which is common given the complexity of tokamak actuators and diagnostics. Control methods must therefore handle partial failures, saturation, and state dropouts. RL approaches explicitly designed for such conditions \citep{fei2020learn}, could play an important role in making controllers more reliable in practice.

% \textbf{Changing Dynamics over time : } Although we leverage more than a decade of DIII-D data, the tokamak has undergone several upgrades to both actuators and diagnostics over this period. As a result, the underlying POMDP dynamics have shifted. This issue is documented in \citet{sonkermulti} and is also evident experimentally, since older discharges cannot be reliably reproduced using identical actuator commands and often require additional adjustments. Such non-stationarity poses a significant challenge for offline RL, which typically assumes fixed dynamics. One pathway forward is to encode configuration-dependent changes directly into the dynamics model through contextual or latent variables \citep{shaj2022hidden}, and another is to adapt the policy online to the dynamics observed at test time \citep{hansen2020self}. Incorporating these techniques into offline RL frameworks offers promising opportunities for robust tokamak control.
\textbf{Changing Dynamics Over Time.}
Our dataset spans more than a decade of DIII-D operation, during which the machine underwent multiple upgrades, causing the underlying dynamics to drift over time. Prior work \citep{sonkermulti} notes that old discharges cannot be replayed with identical commands for this reason. This non-stationarity is difficult for offline RL, which typically assumes fixed dynamics. Bootstrapped models help capture some of this variation, but not fully. Future improvements could include latent context variables for configuration-dependent changes \citep{shaj2022hidden} or online policy adaptation at test time \citep{hansen2020self}.
To conclude, we developed offline RL approaches for rotation-profile control and evaluated several model-free and model-based algorithms. PPO trained with experience from a probabilistic dynamics ensemble produced the best policy, which we then deployed on the DIII-D tokamak. Although real-world testing was limited, the results demonstrate meaningful control of a previously unaddressed profile-regulation task. The probabilistic training strategy shows promising sim2real transfer, and the broader methodology offers an early step toward general-purpose, data-driven profile control in fusion plasmas. 
% \update{Finally, the methodology developed is general purpose and adaptable towards future tokamaks design. Since we do not place any constraints on essential actuators, our method can be applied to any future tokamak design, even non-beam tokamaks \citep{wilson2025heating}.} 
% \update{Finally, because our approach does not impose constraints on specific actuators, it is adaptable to future tokamak designs, including devices without neutral-beam systems \citep{wilson2025heating}.}

\acks{This work would not have been possible without support from many people. We wish to thank the incredible staff at DIII-D that supported us during this experiment. We also thank Aravind Venugopal and Namrata Deka for discussions related to data processing and dynamics modeling, as well as, Yang Fu and Haomin Bao for their help in running offline RL benchmarks. This material is based upon work supported by the U.S. Department of Energy, Office of Science, Office of Fusion Energy Sciences, using the DIII-D National Fusion Facility, a DOE Office of Science user facility, under Awards DE-FC02-04ER54698, DE-SC0024544 and DE-SC0015480.}

\section*{Disclaimer}
This report was prepared as an account of work sponsored
by an agency of the United States Government. Neither
the United States Government nor any agency thereof, nor
any of their employees, makes any warranty, express or implied, or assumes any legal liability or responsibility for the
accuracy, completeness, or usefulness of any information,
apparatus, product, or process disclosed, or represents that
its use would not infringe privately owned rights. Reference
herein to any specific commercial product, process, or service by trade name, trademark, manufacturer, or otherwise
does not necessarily constitute or imply its endorsement, recommendation, or favoring by the United States Government
or any agency thereof. The views and opinions of authors
expressed herein do not necessarily state or reflect those of
the United States Government or any agency thereof.

\bibliography{l4dc2026-sample}

\section{Appendix}

\subsection{RPNN Training}

\subsubsection{State and Actuator Space -}
The dynamics model (an ensemble of RPNN networks) models the transitions on a larger state and action space, in comparison to the policy. This is shown in table \ref{tab:signals}. Variables marked under Actuators and States are used in dynamics modeling. Only a subset of the states is provided as observations to the policy (marked under observations), and a subset of actuators is controlled by the policy (marked under actions).  We do this to simplify the learning process and also because other actuators have different purposes other than controlling the selected task. Note that last action taken is provided in the observation space of the policy.

\begin{table}
\centering
\setlength{\tabcolsep}{6pt}
\renewcommand{\arraystretch}{1.25}

\begin{tabular}{l >{\raggedright\arraybackslash}p{0.24\linewidth} cccc}
% \begin{tabular}{p{0.12\linewidth} >{\raggedright\arraybackslash}p{0.42\linewidth} p{0.11\linewidth} p{0.11\linewidth} p{0.11\linewidth} p{0.11\linewidth}}
\toprule
\textbf{Signal Group} & \textbf{Signals} &
\textbf{Actuator} & \textbf{State} &
\textbf{Action} & \textbf{Observation} \\
\midrule

\multirow{5}{*}{\textbf{Scalar States}}
  & $\beta_{N}$ (Normalized Plasma Pressure) & \xmark & \cmark & \xmark & \xmark \\
  & $l_i$ (Internal Inductance) & \xmark & \cmark & \xmark & \xmark \\
  & Line Averaged Density & \xmark & \cmark & \xmark & \xmark \\
  & Loop Voltage & \xmark & \cmark & \xmark & \xmark \\
  & MHD Stored Energy & \xmark & \cmark & \xmark & \xmark \\
\midrule

\multirow{6}{*}{\textbf{Profile States}}
  & Rotation & \xmark & \cmark & \xmark & \cmark \\
  & Density & \xmark & \cmark & \xmark & \xmark \\
  & Ion Temperature & \xmark & \cmark & \xmark & \xmark \\
  & Electron Temperature & \xmark & \cmark & \xmark & \xmark \\
  & Pressure & \xmark & \cmark & \xmark & \xmark \\
  & Safety Factor $q$ & \xmark & \cmark & \xmark & \xmark \\
  
\midrule

\textbf{Shape Variables}
  & \makecell[l]{Elongation,\\ Upper Triangularity,\\ Bottom Triangularity, \\$a_{\text{minor}}$, \\
  Radial and vertical \\
  positions
  of magnetic axis}
  % Radial position
  % and vertical positions \\
  % of magnetic axis
  & \cmark & \xmark & \xmark & \xmark \\
\midrule

\multirow{2}{*}{\textbf{Neutral Beam Variables}}
  & Power Injected & \cmark & \xmark & \cmark & \cmark \\
  & Torque Injected & \cmark & \xmark & \cmark & \cmark \\
\midrule

\textbf{Gas Puffing}
  & GasA voltage
  & \cmark & \xmark & \cmark & \cmark \\
\midrule

\textbf{Electron Cyclotron Heating}
  & ECH Total Power
  & \cmark & \xmark & \cmark & \cmark \\
\midrule

\textbf{Other Actuators}
  & \makecell[l]{Current Target,\\ Toroidal Field}
  & \cmark & \xmark & \xmark & \xmark \\
\midrule

\textbf{Targets}
  & \makecell[l]{Rotation Target (t), \\ Rotation Target (t+10), \\ Current Error Terms}
  & \xmark & \xmark & \xmark & \cmark \\
\midrule

\multicolumn{2}{l}{\textbf{Total Dimensions}} &
\textbf{12D} & \textbf{25D} & \textbf{4D} & \textbf{20D} \\
\bottomrule
\end{tabular}

\caption{Plasma signals and how they are used as State and Actuators for dynamics modelling. Policy Observation and action space variables are also shown.}
\label{tab:signals}
\end{table}

\subsubsection{Network Architecture and Training details - }

\textbf{Network Architecture : }
\begin{itemize}
    \item \textbf{Encoder:}
    \begin{itemize}
        \item Fully Connected (FC) layer: \texttt{input\_dim} $\times$ 512
        \item FC layer: 512 $\times$ 512
    \end{itemize}
    
    \item \textbf{Memory Unit:}
    \begin{itemize}
        \item Gated Recurrent Unit (GRU) block: 512 $\times$ 256
    \end{itemize}

    \item \textbf{Decoder (with residual connections between FC layers):}
    \begin{itemize}
        \item FC layer: 256 $\times$ 512
        \item FC layers: 512 $\times$ 512 (repeated 8 times)
        \item FC layer: 512 $\times$ 128
    \end{itemize}

    \item \textbf{Output Heads:}
    \begin{itemize}
        \item Mean head: 128 $\times$ \texttt{output\_dim}
        \item Log-variance head: 128 $\times$ \texttt{output\_dim}
    \end{itemize}
\end{itemize}

The network predicts the parameters of a probability distribution, and is trained in two stages. The First stage correponds to training with Mean Square error loss (MSE) and the second stage correponds to training only the log variance head with negative log likelihood loss. We use the Adam optimizer with a learning rate of $3 \times 10^{-4}$ and a weight decay of $1 \times 10^{-3}$ in both stages. A 25 model ensemble is trained by bootstrapping on the dataset. We train for 1000 epochs with early stopping (patience = 250 epochs) based on performance on a validation set comprising 10\% of the total data.

\subsection{Offline RL Benchmarking}

Our complete set of results for all algorithms tested are provided in table \ref{tab:rotation-tracking-full}. This includes RMSE tracking at various points along the profile from the core $\psi_n=0$ to the edge $\psi_n=1$ of the plasma.

\begin{table}[htbp]
\centering
\caption{Performance of Offline RL algorithms on Rotation profile tracking error. RMSE values (with standard error in parenthesis) are shown for the whole profile tracking and also across $\psi_n$ values which vary from 0 (core) to 1 (edge) of plasma profile (lower is better). These results were computed across shots 161409, 161410, 161412 in simulation with 10 seeds across each shot.}
\label{tab:rotation-tracking-full}
\setlength{\tabcolsep}{3pt}  % reduce horizontal padding
\resizebox{\linewidth}{!}{
\begin{tabular}{lccccccc}
\toprule
Algorithm & Profile & $\psi_n = 0.09$ & $\psi_n = 0.18$ & $\psi_n = 0.39$ & $\psi_n = 0.58$ & $\psi_n = 0.79$ & $\psi_n = 0.88$ \\
\midrule
TD3BC & 58.73 (0.51) & 62.64 (3.17) & 57.54 (2.62) & 49.05 (2.11) & 43.73 (1.85) & 39.36 (1.39) & 24.85 (0.39) \\
\textbf{PPO} & 29.5 (0.46) & 45.77 (2.17) & 39.05 (1.42) & 30.16 (0.77) & 26.84 (0.67) & 26.19 (0.62) & 21.12 (0.29) \\
GCIL & 36.48 (0.37) & 50.15 (2.09) & 43.8 (1.87) & 35.32 (1.38) & 31.86 (1.06) & 30.71 (0.83) & 23.47 (0.26) \\
% MPPI & 24.09 (0.34) & 40.49 (1.86) & 35.42 (1.52) & 29.27 (1.14) & 25.89 (0.91) & 23.71 (0.69) & 18.52 (0.43) \\
CQL & 80.79 (0.25) & 73.83 (1.13) & 67.58 (0.81) & 57.14 (1.03) & 50.84 (0.97) & 46.04 (0.59) & 29.29 (0.21) \\
EDAC & 32.07 (0.66) & 47.09 (3.85) & 41.58 (2.8) & 34.2 (1.65) & 30.29 (1.3) & 27.39 (1.32) & 19.31 (0.62) \\
MCQ & 44.64 (0.44) & 55.89 (2.59) & 49.94 (2.53) & 40.48 (2.02) & 35.67 (1.45) & 33.1 (1.07) & 23.11 (0.42) \\
COMBO & 51.46 (0.62) & 59.96 (3.57) & 53.09 (2.6) & 43.05 (1.55) & 38.49 (1.26) & 36.07 (1.3) & 24.96 (0.72) \\
IQL & 54.52 (0.78) & 61.46 (5.43) & 55.12 (4.94) & 45.0 (3.24) & 39.98 (2.2) & 37.69 (1.79) & 26.84 (0.91) \\
MOPO & 84.63 (0.72) & 76.81 (4.38) & 68.82 (3.57) & 56.15 (2.37) & 49.78 (1.84) & 46.42 (1.7) & 31.93 (1.02) \\
MOBILE & 32.99 (0.4) & 45.64 (2.67) & 43.0 (1.79) & 38.02 (0.98) & 33.77 (0.77) & 29.82 (0.55) & 20.13 (0.28) \\
BAMCTS & 57.43 (0.71) & 62.69 (4.0) & 56.27 (2.93) & 46.74 (2.12) & 41.86 (1.91) & 38.61 (1.64) & 25.63 (0.54) \\
RAMBO & 40.05 (0.43) & 51.3 (2.52) & 45.99 (2.03) & 39.14 (1.49) & 36.14 (1.33) & 34.28 (1.26) & 24.43 (0.62) \\
ROMBRL & 91.44 (0.7) & 79.59 (4.18) & 72.05 (3.26) & 59.71 (2.16) & 52.56 (1.7) & 47.16 (1.38) & 28.88 (0.57) \\
\bottomrule
\end{tabular}}
\end{table}

\section{Profile Views}
\begin{figure}[H]
  \centering
  \includegraphics[width=0.8\linewidth]{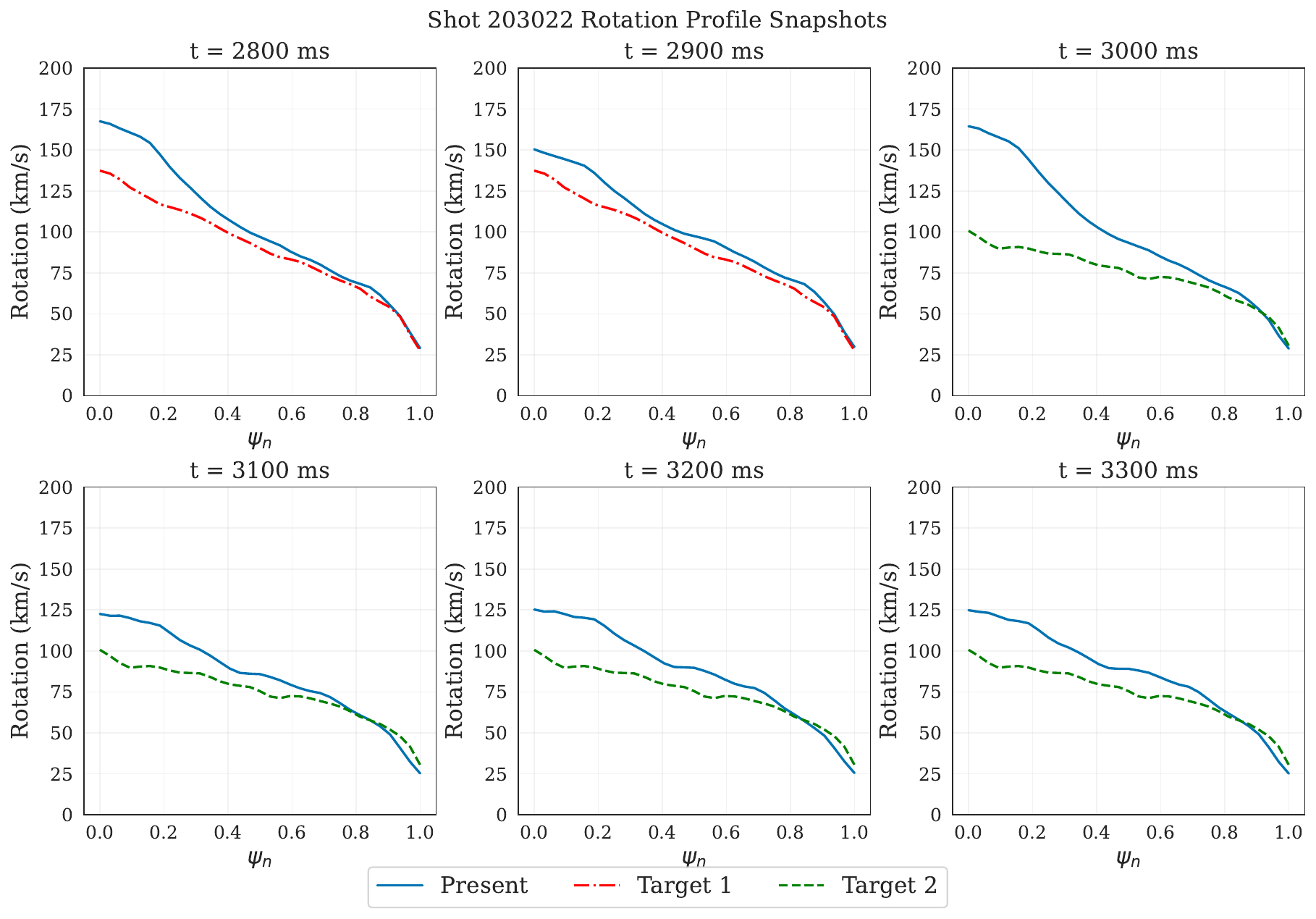}
  \caption{Full RTCAKENN Rotation profile at various time slices during the first target change. The change in target line (red to green) occurs at t=3s. The real rotation profile (blue) starts moving down to get closer to the updated target.}
  \label{fig:full_prof1}
\end{figure}

\begin{figure}[H]
  \centering
  \vspace{-1em}
  \includegraphics[width=\linewidth]{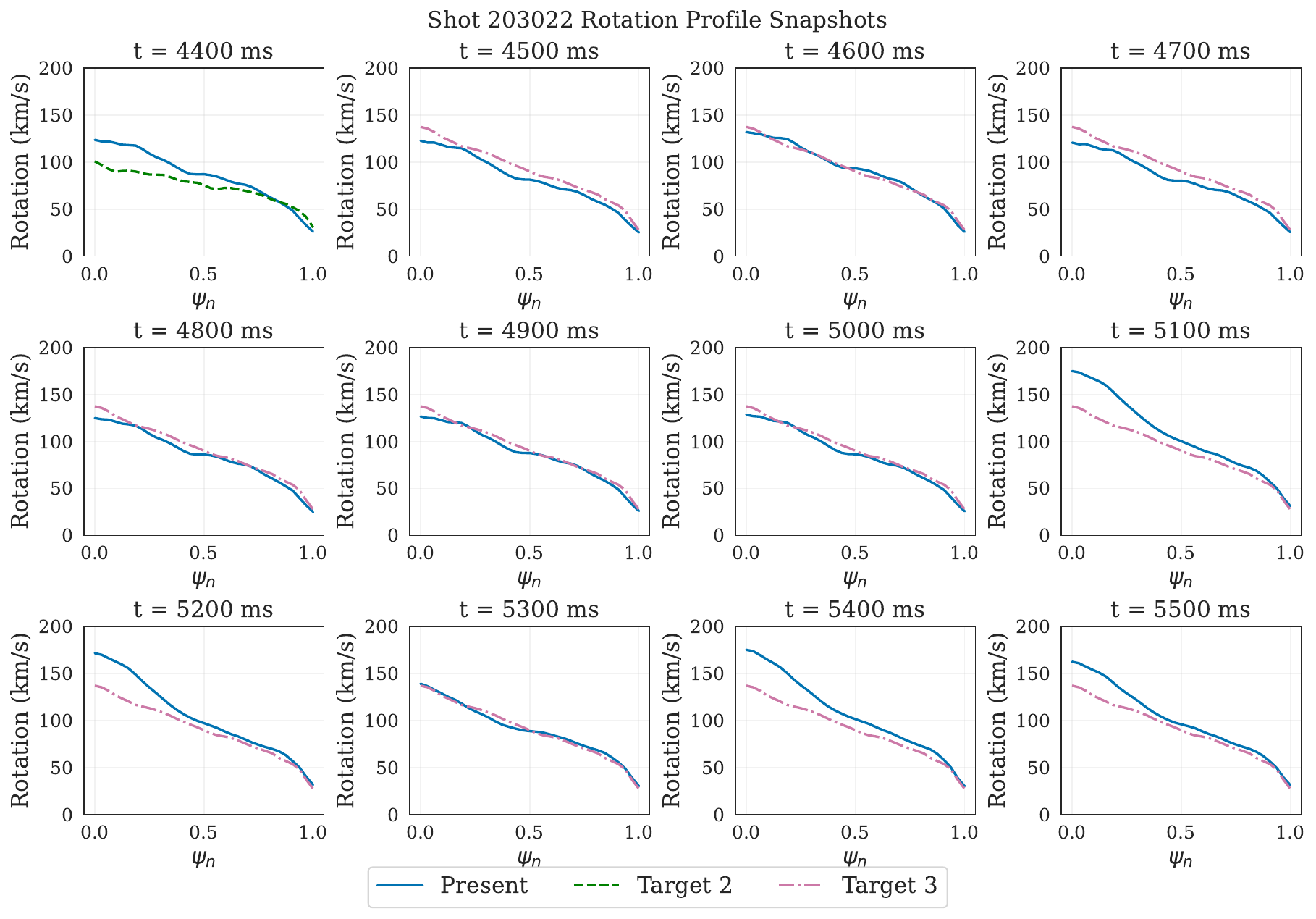}
  \caption{Full RTCAKENN rotation profiles at various time slices during the second target change. The change in target line (green to purple) occurs at t=4.5s. This new target matches the present profile well, hence drastic changes in actuators are not expected. A sudden rise in core rotation is observed, which later drops. This is attributed to either unknown plasma effects or noise in signals.}
  \label{fig:full_prof2}
\end{figure}

In this section, we present full rotation profile views across time. Fig.\ref{fig:full_prof1} shows the profile variation when the first target change occurs. This target change corresponds to a decrease in profile at t = 3s. Fig.\ref{fig:full_prof2} shows the profile changes that occur during the second target change at t = 4.5 s. At this point, more variations occur in the plasma, which can be observed in the figure. The rise in core rotation followed by a drop is attributed to unknown plasma effects. We examined the presence of tearing mode activity and large sawtooth instabilities, however, no clear evidence of either was detected.

% \vspace{-2em}

\section{Gas Flow}
\begin{figure}
  \centering
  % \vspace{-1em}
\includegraphics[width=0.8\linewidth]{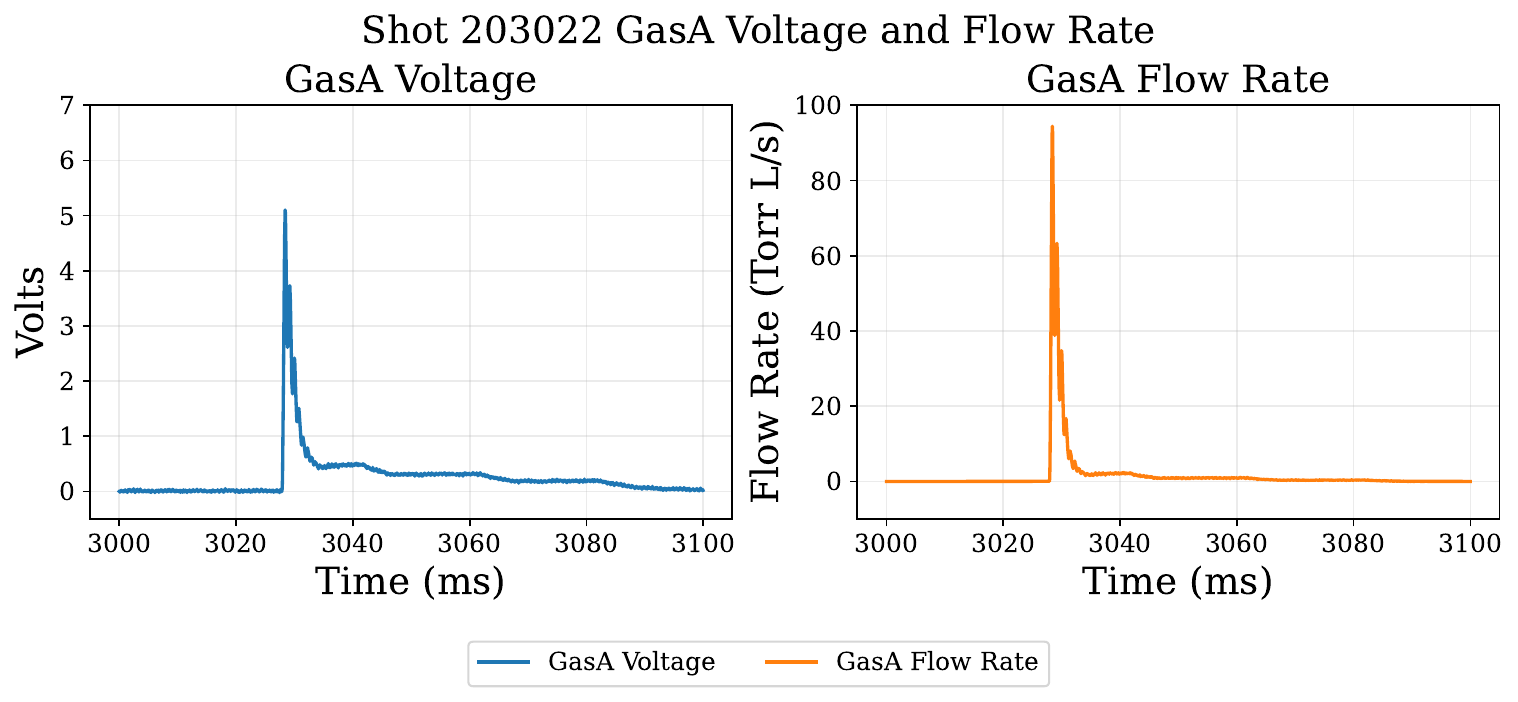}
  \caption{Gas Voltage and the corresponding gas flow rate increase at approx t=3s, corresponding to the target change.}
  \label{fig:gas_flow}
\end{figure}

In this section, plots showing the gas flow rate into the plasma are shown in Fig.~\ref{fig:gas_flow}. These plots show the actual amount of gas flowing into the plasma, corresponding well to the gas valve voltage, which the policy controls.

% \clearpage % (removing this avoids forcing a final blank page)
%
% If you still get a trailing blank page, it is usually due to a final implicit
% \clearpage executed by \end{document} after a float-only last page.
% This disables that final \clearpage (safe if all floats are already placed).
% \makeatletter\let\clearpage\relax\makeatother
% \clearpage
\end{document}